\documentclass[letterpaper, 10 pt, conference]{ieeeconf}

\IEEEoverridecommandlockouts
\usepackage{cite}

\usepackage[pagewise]{lineno}
\usepackage{amsmath}
\usepackage{amssymb}
\usepackage{wrapfig}
\usepackage{pifont}
\usepackage{tabularray}

\usepackage{multirow} 
\usepackage{hyperref}
\usepackage{mathtools}
\usepackage{bm}
\usepackage{algorithm}
\usepackage{algpseudocode}

\usepackage{graphicx}
\usepackage{subcaption}
\usepackage{xcolor}
\usepackage{booktabs}

\graphicspath{{./figures/}}

\title{\LARGE \bf
Disturbance Observer-based Control Barrier Functions with Residual Model Learning for Safe Reinforcement Learning
}
\author{Dvij Kalaria$^{1}$, Qin Lin$^{2}$, and John M. Dolan$^{1}$
\thanks{$^{1}$The authors are with the Robotics Institute, Carnegie Mellon University {\tt\small dkalaria@andrew.cmu.edu},{\tt\small jdolan@andrew.cmu.edu}}
\thanks{$^{2}$Qin Lin is with the Engineering Technology Department, University of Houston {\tt\small qlin21@central.uh.edu}}}

\begin{document}

\maketitle
\thispagestyle{empty}
\pagestyle{empty}


\begin{abstract}

Reinforcement learning (RL) agents need to explore their environment to learn optimal behaviors and achieve maximum rewards. However, exploration can be risky when training RL directly on real systems, while simulation-based training introduces the tricky issue of the sim-to-real gap. Recent approaches have leveraged safety filters, such as control barrier functions (CBFs), to penalize unsafe actions during RL training. However, the strong safety guarantees of CBFs rely on a precise dynamic model. In practice, uncertainties always exist, including internal disturbances from the errors of dynamics and external disturbances such as wind. In this work, we propose a new safe RL framework based on disturbance rejection-guarded learning, which allows for an almost model-free RL with an assumed but not necessarily precise nominal dynamic model. We demonstrate our results on the Safety-gym benchmark for Point and Car robots on all tasks where we can outperform state-of-the-art approaches that use only residual model learning or a disturbance observer (DOB). We further validate the efficacy of our framework using a physical F1/10 racing car. Videos: https://sites.google.com/view/res-dob-cbf-rl
\end{abstract}


\setlength{\textfloatsep}{12pt plus 0.0pt minus 2.0pt}

\section{Introduction}

Recently, reinforcement learning (RL) has been successful in solving various challenging control tasks on real systems, such as legged locomotion \cite{Kumar2021RMARM}, manipulation \cite{Morales2021ASO}, and autonomous driving \cite{Kiran2020DeepRL}. Most of these approaches use sim-to-real transfer, deploying policies trained on simulators with varying model parameters to mimic real-world uncertainty, and then applying the same policy to the real robot in the hope of achieving robustness to real-world disturbances. Some very recent works have moved towards performing RL training directly on real systems \cite{Wu2022DayDreamerWM}. 
Such systems should remain safe during training \cite{Paduraru2021ChallengesOR} to avoid costly damage, as obstacle collisions that inform the policy gradient to enforce safety can be very expensive. To address this issue fundamentally, safe RL algorithms that respect safety while learning have been developed in recent years. There may be different definitions of safety, as discussed in \cite{Garca2015ACS,Brunke2021SafeLI}, but we focus on satisfying hard safety constraints, such as avoiding certain forbidden states, as our safety specifications.



\begin{figure}
    \centering
    \includegraphics[width=.45\textwidth]{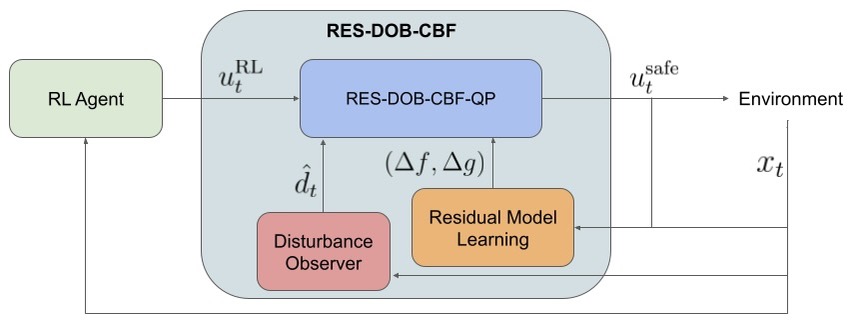}
    \caption{Propose safe RL using RES-DOB-CBF. At each time step $t$, the RL agent’s action $u_{t}^{\mathrm{RL}}$ is monitored and possibly overridden by the RES-DOB-CBF-based safety filter to avoid any unsafe actions.
    \label{fig:overview}}
\end{figure}

Among different safe RL paradigms, the most common approach is to use safety filters like control barrier functions (CBFs) \cite{cheng2019end} \cite{9147584} on the actions generated by RL agents, overriding them whenever necessary to satisfy the constraints. An advantage of this design is that the safety filter can be integrated as a plug-in into any RL algorithm with minimal modification. However, the disadvantage is that these safety filters require precise dynamic models to ensure strong safety guarantees. Using a known model works well for a simulator with known dynamics, but is challenging to deploy on a real robot on certain tasks where it would be difficult to do model identification. Also, the robot model can change with time for certain agile tasks such as racing \cite{Kalaria2023AdaptivePA}, offroad driving \cite{Wang2023PayAT}, drones \cite{Barawkar2022AdaptiveCO}, etc. 

Consequently, most existing learning-for-control frameworks assume a nominal model, using a discrete dynamic model, such as $\hat{F}(x_k, u_k)$. To quantify the difference from the true model, the state at the next time step, $x_{k+1} = F(x_k, u_k)$, is used as a label to represent the true model. This process is referred to as \emph{residual model learning} in our paper.  
The residual model learning can handle limited model uncertainties very well but usually fails to compensate for fast-changing external disturbances, such as wind, that influence the robot’s dynamics. When the learned model is poor—e.g., if the machine learning has not converged or suffers from sampling inefficiency—the safety filter based on the sum of the nominal model and the learned discrepancy may still fail to provide a safety guarantee. 

In \cite{Cheng2022SafeAE}, the authors claim to use a simple nominal model for efficient RL training without the need for a high-fidelity model. Instead, they employ a Disturbance Observer (DOB) to compensate for disturbances, including model errors. As a result, the RL approach is \emph{almost model-free}, eliminating the need for continuous learning of the dynamic model, since the model discrepancy is addressed by the DOB, which provides stronger guarantees for uncertainty quantification. However, when the nominal model has a significant discrepancy from the true model and there are also external disturbances, the DOB must compensate for all of these factors, which can lead to slow convergence of the estimation.



Our previous work \cite{zhang2024disturbance} proposes a disturbance rejection-guarded learning (DRGL) approach that combines a disturbance observer with residual model learning. In the forward path of DRGL, residual model learning provides prior knowledge of the disturbance to the disturbance observer, leading to faster estimation convergence. In the backward path, the disturbance observer compensates for errors in residual model learning, adding an extra layer of assurance. In this work, we extend this idea into safe RL. Refer to Fig. \ref{fig:overview} for an overview of our approach, where RES-DOB-CBF is a robust safety filter that overrides unsafe actions under uncertainty during training. The filter consists of a disturbance observer that compensates for rapidly changing continuous external disturbances, and a residual model learning component that addresses imperfections in the nominal model due to incorrect parameters or complex physics.

Our work differs from \cite{zhang2024disturbance} in several ways: (1) The nominal dynamic model considered in \cite{zhang2024disturbance} is linear, whereas our work addresses a nonlinear dynamic model; (2) The control task in \cite{zhang2024disturbance} is a standard tracking problem without safety constraints, while our work deals with an optimal control problem that includes safety specifications; (3) \cite{zhang2024disturbance} focuses solely on supervised learning for disturbance, while our work involves reinforcement learning for a control policy that is robust to disturbances. 

The contributions of our work are as follows:

\begin{enumerate}
\item We propose an almost model-free safe RL framework that relies only on a nominal model and can optimize a safe control policy in the presence of internal and external disturbances. \item The disturbance-rejection guarded learning component integrates residual model learning and DOB to leverage the advantages of both. This integration is expected to provide stronger convergence guarantees than using each component alone. Additionally, this component facilitates optimization using a robust CBF, which outputs provably safe control actions in the presence of uncertainty.
\item We comprehensively validate our approach on a modified Safety Gym benchmark \cite{safetygym}, where we outperform vanilla PPO\_Lagrangian, DOB+CBF \cite{Cheng2022SafeAE}, and Residual Model+CBF approaches on all tasks. Hardware experiments with a physical F1/tenth racing car are also conducted to validate the efficacy of our framework.
\end{enumerate}

\section{Methodology} \label{sec:methodology}

We consider a general control-affine nonlinear system: 
\begin{equation} \label{eq:env_model}
    \dot{{x}}(t) = f({x}(t)) + g({x}(t)) {u}(t) + {d}(t)
\end{equation}
where ${x}(t) \in \mathcal{X} \subset \mathbb{R}^n$ is the state vector, ${u}(t) \in \mathcal{U} \subset \mathbb{R}^m$ is the input vector, $f:\mathbb{R}^n \rightarrow \mathbb{R}^n$ and $g:\mathbb{R}^n \rightarrow \mathbb{R}^n$ are the Lipschitz-continuous model functions, and ${d}(t) \in \mathcal{D} \subset R^n$ represents nonlinear time-varying disturbance.  

\subsection{Model Residual learning} \label{sec:model_res_learning}

Moving back to \eqref{eq:env_model}
, assume that we do not have an accurate model for $f$ and $g$, but only a nominal model $\hat{f}$ and $\hat{g}$. The difference between the actual models $f$ and $g$ and the nominal models $\hat{f}$ and $\hat{g}$ arises from model errors due to system identification or from dynamics that are difficult to model, such as the environment affecting the tires of a car. The nominal model can also be a low-fidelity model, such as the kinematic bicycle model of a car, rather than a high-fidelity dynamic model. Hence, we can modify \eqref{eq:env_model} to:

\begin{equation}
\begin{aligned}
    \dot{{x}}(t) &= \hat{f}({x}(t)) + \hat{g}({x}(t)) {u}(t) \\
    &+ {\Delta f}({x}(t),\theta_f)+{\Delta g}({x}(t),\theta_g)u  
\end{aligned}
\end{equation}
where $\theta$ is from by a feed-forward neural network which takes the state, ${x}$, and outputs the model residual represented by ${\Delta f}({x}(t),\theta_f)+{\Delta g}({x}(t),\theta_g) u$. We estimate $\theta$ from online collected data by optimizing the following loss function

\begin{equation} \label{eq:res_model_opt}
\begin{aligned}
\theta^{*}=\underset{\theta \in \Theta}{\operatorname{argmin}} &\mathcal{L}({\Delta f}({x}(t),\theta_f)+{\Delta g}({x}(t),\theta_g) \dot{x} \\
&- \hat{f}(x) - \hat{g}(x) u) 
\end{aligned}
\end{equation}

To solve for an approximate solution, we can update $\theta_f, \theta_g$ iteratively with each new collected data element with the following update rule with $\gamma$ as the learning rate:

\begin{equation}
\theta_{f/g,k+1} = \theta_{f/g,k} - \gamma \frac{\delta \mathcal{L}_k}{\delta \theta_{f/g,k}} 
\label{eq:update_rule}
\end{equation}

\subsection{Disturbance Observer (DOB)} \label{sec:dob}

DOB has been widely used in control theory for uncertain systems \cite{Chen2016DisturbanceObserverBasedCA}. The idea is to address all unknown uncertainties from internal disturbances (model errors) and external disturbances as lumped disturbances. In this work, we use the DOB formulation in \cite{Cheng2022SafeAE}. DOB consists of two components: a state predictor and a piecewise-constant (PC) estimation law. The state predictor is given by:

\begin{equation} \label{eqn:dob1}
\begin{aligned}
\dot{\hat{{x}}}(t)&=f({x}(t))+g({x}(t)) {u}(t)+{\Delta f}({x}(t),\theta_f) \\
&+{\Delta g}({x}(t),\theta_g)u + \hat{{d}}(t)-a \tilde{{x}}(t)
\end{aligned}
\end{equation}
where $\tilde{{x}} = \hat{{x}} - {x}$ denotes the state prediction error, $a > 0$ is a constant, and $\hat{{d}}(t)$ is the estimated disturbance. The disturbance estimation is updated according to

\begin{equation} \label{eqn:dob2}
\left\{\begin{aligned}
\hat{{d}}(t) & =\hat{{d}}(i T), \quad t \in[i T,(i+1) T), \\
\hat{{d}}(i T) & =-\frac{a}{e^{a T}-1} \tilde{{x}}(i T), i=0,1, \ldots,
\end{aligned}\right.
\end{equation}
where $T$ is the estimation sampling time. Under the assumption that the unknown disturbance is locally Lipschitz continuous with a known bound, the estimation error from DOB, estimation error $||\hat{{d}}(t) - {d}(t)||$ can be bounded by a pre-computed error bound \cite{Cheng2022SafeAE}.

\subsection{Residual model and DOB-based Control Barrier Function (RES-DOB-CBF)} \label{subsec:res_dob_cbf}

CBF is a popular model-based tool to enforce constraints with a known dynamic model and safety specifications. We define a continuous and differentiable safety function $h({x}): \mathcal{X} \xrightarrow{} \mathbb{R}$. The \emph{superlevel set} $\mathcal{C} \in \mathbb{R}^n$ can be named as a safe set. Let the set $\mathcal{C}$ obey $\mathcal{C} = \{{x} \in \mathcal{X}: h({x})\geq 0 \}$

The safety function $h({x})$ is called a CBF of a control affine system defined in Eq. \eqref{eq:env_model} with relative degree of one, if there exists a $\beta >0$ such that

\begin{equation}
    \underset{{u} \in \mathcal{U}}{\sup}\left[\mathcal{L}_f h({x}) + \mathcal{L}_g h({x}) {u} + \beta (h({x}))\right] \geq 0
    \label{cbf}
\end{equation}
for all ${x} \in \mathcal{X}$. $\mathcal{L}_fh({x})$ and $\mathcal{L}_gh({x})$ are Lie derivatives. $\beta(h({x}))$ is particularly chosen as a special class $\mathcal{K}$ function $\beta h({x})$. The solution ${u}$ assures that the set $\mathcal{C}$ is a forward invariant, i.e., any solution starting at any $x(0) \in \mathcal{C}$ satisfies ${x}(t) \in \mathcal{C}$ for $\forall t \ge 0$. For a higher-order CBF, i.e., if $L_gh({x})=0$ and we have $m$ such that $L_gL_f^{m-1}({x}) \neq 0$ and $L_gL_f^{m-2}({x}) = 0$

\begin{equation}
\sup _{u \in \mathcal{U}} \mathcal{L}_{f}^{m} h(x)+\mathcal{L}_{g} \mathcal{L}_{f}^{m-1}h(x)u+O(h(x))+\beta_{m}\left(\phi_{m-1}(x)\right) \geq 0
\end{equation}
where $L_{f}^{m} h(x)=\frac{\partial L_{f}^{m-1} h(x)}{\partial x} f(x), \mathcal{L}_{g} \mathcal{L}_{f}^{m-1} h(x)=\frac{\partial L_{f}^{m-1} h(x)}{\partial x} g(x)$ and $\left[\mathcal{L}_{f}^{m-1} h(x)\right]_{x}=\frac{\partial L_{f}^{m-1} h(x)}{\partial x}$, $O(h(x))=\sum_{i=1}^{m-1} \mathcal{L}_{f}^{i}\left(\beta_{m-i} \circ \phi_{m-i-1}\right)(x)$, and $\phi_{i}(x)=\dot{\phi}_{i-1}(x)+\beta_{i}\left(\phi_{i-1}(x)\right), \phi_{0}=h(x)$.

We will use a robust CBF with an arbitrary relative degree:

\begin{equation}
\sup _{u \in \mathcal{U}} \mathcal{K}(t, x, u)+\beta_{m}\left(\phi_{m-1}(x)\right) \geq 0
\end{equation}
where $\mathcal{K}(t, x, u) \triangleq L_{f}^{m} h(x)+L_{g} L_{f}^{m-1} h(x) u+O(h(x))+\left[\mathcal{L}_{f}^{m-1} h(x)\right]_{x} \hat{d}(t)$. ${f}={\hat{f}}+{\Delta f}$ and ${g}={\hat{g}}+{\Delta g}$ account for residual model learning, with $\hat{d}$ representing the estimated disturbance. This way, we incorporate uncertainty compensation from both residual learning and the disturbance observer.




\subsection{Safe-RL policy training with RES-DOB-CBFs} \label{sec:saferl}


To obtain a safe control, we solve the following quadratic programming (QP) problem based on the RES-DOB-CBF :

\begin{equation} \label{eq:filter_opt}
\begin{aligned}
u_{\text {safe }} & =\underset{u \in \mathcal{U}}{\operatorname{argmin}} \frac{1}{2}\left(u-u_{\mathrm{RL}}\right)^{T} P\left(u-u_{\mathrm{RL}}\right)\\
\text { s.t. } & \mathcal{K}(t, x, u)+\beta_{m}\left(\phi_{m-1}(x)\right) \geq 0
\end{aligned}
\end{equation}
where $P$ is the positive-definite weighting matrix, $u_{\text{RL}}$ is the nominal control generated by the RL agent, $u_{\text{safe}}$ is the final safe control. To have an efficient and safe RL controller, it is critical to obtain a safety filter based on an accurate dynamic model. If the safety filter is inaccurate, it will not guard the agent sufficiently, potentially leading to violations of safety constraints. Conversely, if the safety filter is overly conservative, it will intervene unnecessarily with the actions generated by the RL, hindering the learning of an agile policy. Thanks to the uncertainty compensation from our DOB and residual model learning, we provide robust constraints in \eqref{eq:filter_opt} for an accurate but non-conservative safety filter. 
To have an efficient and safe RL controller, it is critical to use a safety filter based on an accurate dynamic model. If the safety filter is inaccurate, it will not guard the agent sufficiently, leading to potential violations of safety constraints. Conversely, if the safety filter is too conservative, it will unnecessarily intervene with the actions generated by the RL algorithm, preventing the agent from learning an agile policy. The pseudo-code for the overall flowchart is described in Algorithm \ref{alg:safe_rl}. 


\begin{algorithm}
\caption{RES-DOB-CBF-based safe RL}
\label{alg:safe_rl}
\begin{algorithmic}[1]
\Require Initial policy $\pi_{\theta}$, no. of episodes $N$, no. of steps per episode $M$, no. of policy updates $G$, nominal dynamics $\dot{x}=\hat{{f}}(x)+\hat{{g}}(x) u$, DOB's estimation error bound $\gamma$
\For{$i = 1$ to $N$}
    \For{$t = 1$ to $M$}
        \State Obtain action $u_{t}^{\mathrm{RL}}$ from policy $\pi_{\theta}$
        \State Obtain disturbance estimation $\hat{d}_{t}$ from the DOB defined via Eq. \eqref{eqn:dob2}
        \State Obtain safe action $u_{t}^{\text {safe}}$ from RES-DOB-CBF-QP defined in Eq. \eqref{eq:filter_opt}, using $u_{t}^{\mathrm{RL}}, \gamma$, and $\hat{d}_{t}$
        \State Execute action $u_{t}^{\text {safe}}$ in the environment
        \State Update residual model weights $\theta_f$ and $\theta_g$ as defined in Eqs. \eqref{eq:res_model_opt}-\eqref{eq:update_rule}
        \State Add transition $\left(x_{t}, u_{t}^{\text {safe}}, x_{t+1}, r_{t}\right)$ to replay buffer $\mathcal{D}$
        \For{$j = 1$ to $G$}
            \State Sample mini-batch $\mathcal{B}$ from $\mathcal{D}$
            \State Update policy $\pi_{\theta}$ using $\mathcal{B}$
        \EndFor
    \EndFor
\EndFor
\end{algorithmic}
\end{algorithm}
In each step of each episode, we first generate a nominal control action from the existing RL controller (Line 3). We then estimate the disturbance (Line 4) and solve the robust QP-CBF optimization to obtain a safe control action (Line 5), which is the final action executed. We also use the newly collected state and control to update the residual model represented by neural networks (Line 7). Note that the safe control action $u_{t}^{\text{safe}}$, rather than $u_{t}^{\mathrm{RL}}$, is stored in the replay buffer (Line 8) for a safe policy update (Lines 9-12).

\section{Experimental Results} \label{sec:results}
In this section, we present our experimental results from both simulation and physical testbed environments. We compare our approach (called \emph{RES-DOB-CBF}) with the following RL algorithms: (1) without CBF, (2) with CBF but without any model error compensation, (3) with CBF + DOB (called \emph{DOB-CBF}) \cite{Cheng2022SafeAE}, and (4) with CBF + residual model learning (\emph{RES-CBF}) \cite{Munoz2022}. All videos are available at \href{https://sites.google.com/view/res-dob-cbf-rl}{https://sites.google.com/view/res-dob-cbf-rl}.

\begin{figure}[htbp]
\centering
\begin{subfigure}{.15\textwidth}
    \centering
    \includegraphics[width=\textwidth]{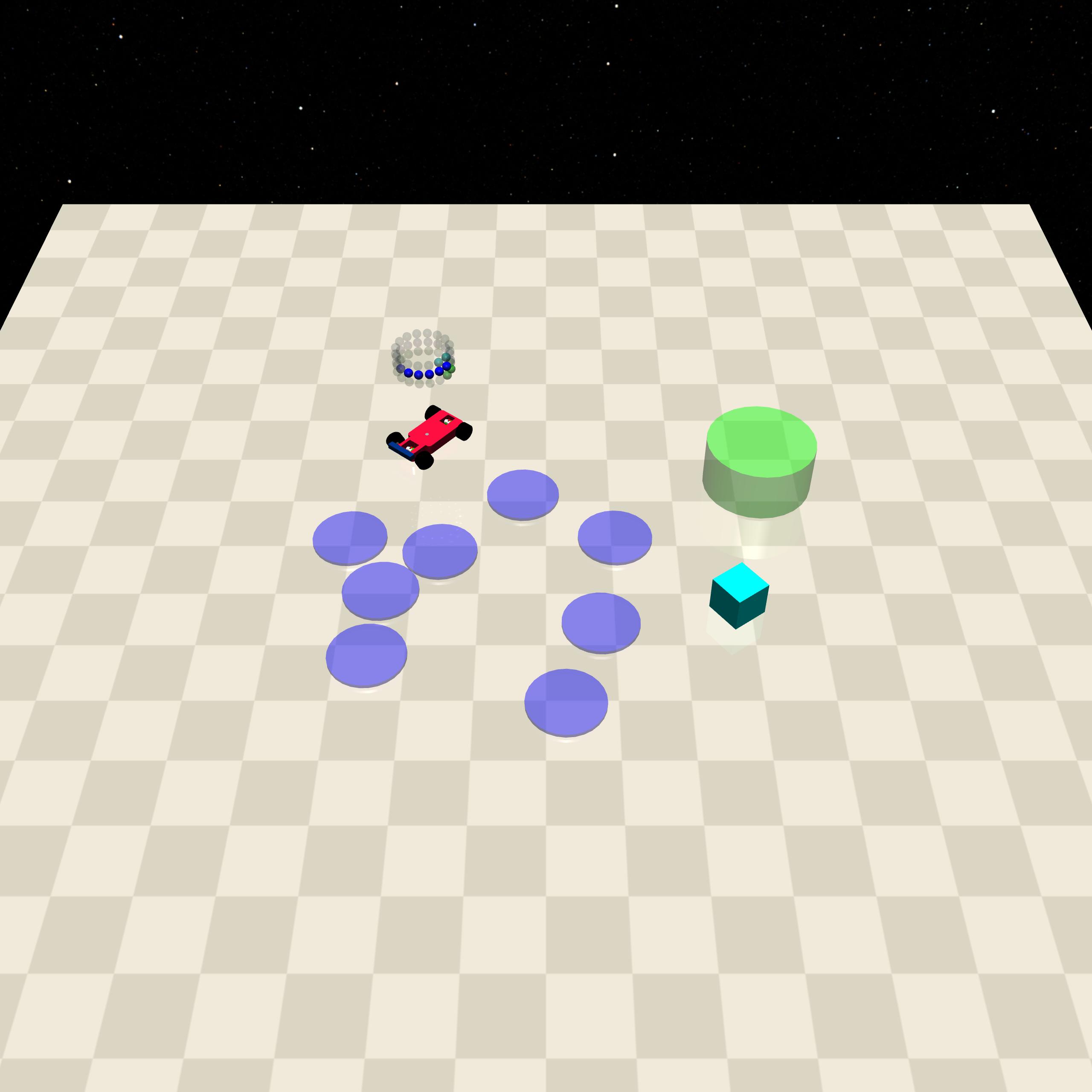}
    \caption{Goal Task}
\end{subfigure}
\begin{subfigure}{.15\textwidth}
    \centering
    \includegraphics[width=\textwidth]{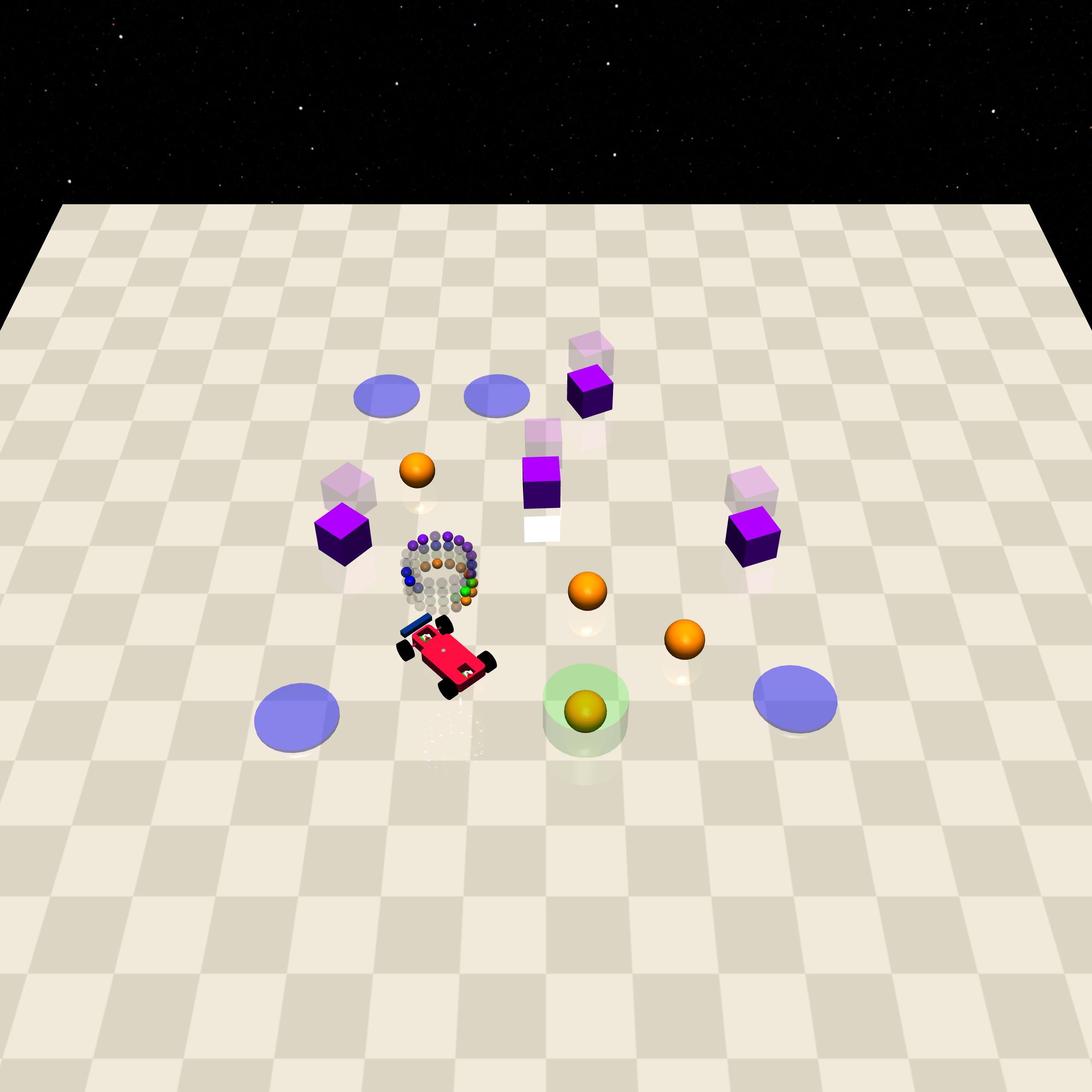}
    \caption{Button}
\end{subfigure}
\begin{subfigure}{.15\textwidth}
    \centering
    \includegraphics[width=\textwidth]{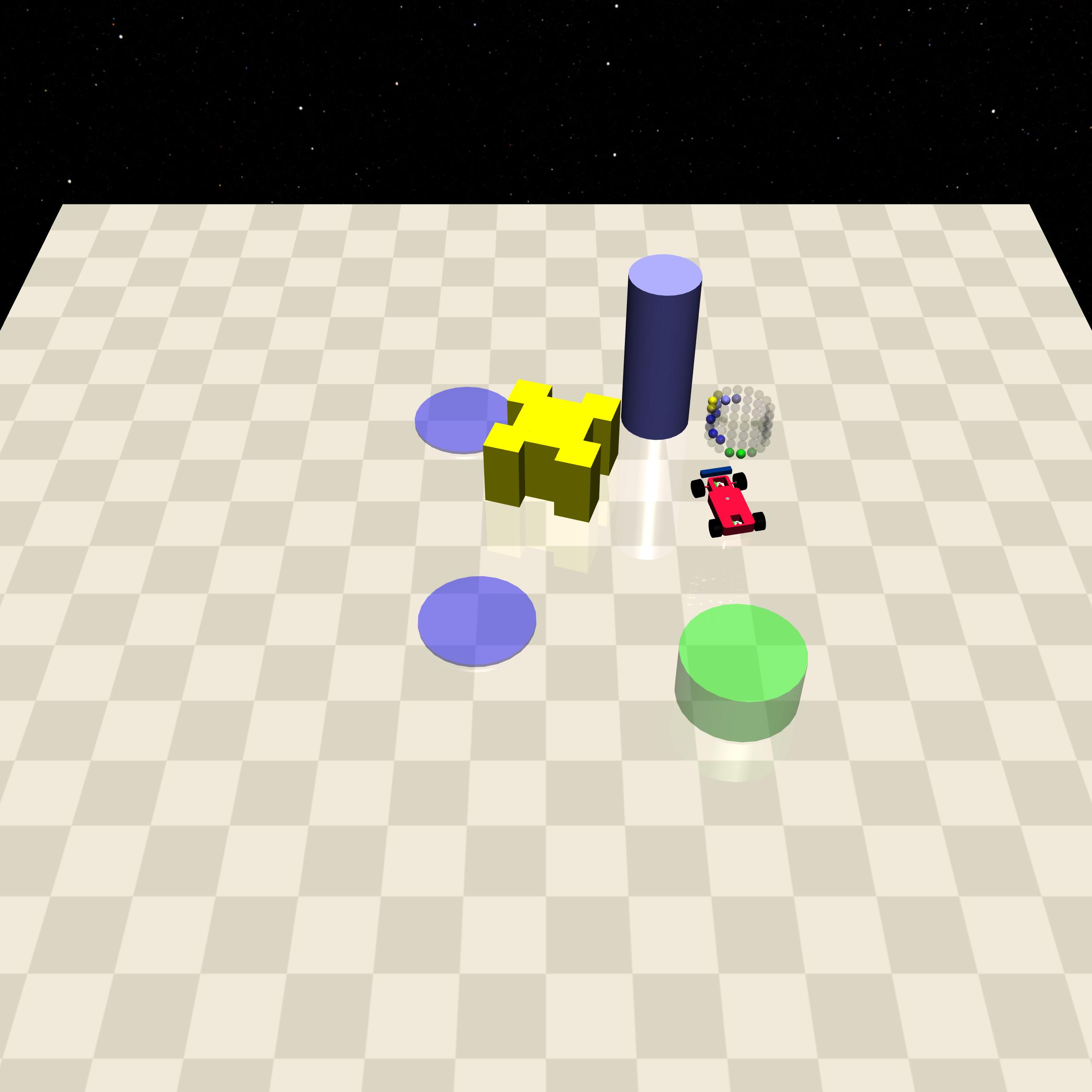}
    \caption{Push}
\end{subfigure}
\caption{Safety-gym environment tasks}
\label{fig:safety_gym_eg}
\end{figure}

We first perform experiments on the Safety-gym benchmark \cite{Achiam2019BenchmarkingSE} using the Point and Car robots. The environment selections include Goal1, Goal2, Button1, Button2, Push1, and Push2, see Fig. \ref{fig:safety_gym_eg}. The obstacles consist of hazards, pillars, vases, and gremlins. Safety-gym is a well-known benchmark for evaluating the performance of Safe RL algorithms. We use the \emph{PPO\_Lagrangian} algorithm \cite{Achiam2019BenchmarkingSE}, which regulates the negative reward received from violating constraints to achieve a target cost. We set the target cost to 25 for all experiments, as in \cite{Achiam2019BenchmarkingSE}, meaning the agent is allowed to violate constraints for up to 25 time steps in an episode of length 2000. To validate our approach’s efficacy in handling uncertainties, we include \textbf{internal disturbances caused by model uncertainty, e.g., 50$\%$ model errors in the kinematic models}. We also add \textbf{wind as an external disturbance} with a fixed magnitude, 0.25 m/s\textsuperscript{2} for the Point robot and 2.5 m/s\textsuperscript{2} for the Car robot (equivalent acceleration generating force) and a continually changing direction with a constant angular rate of $5$ Hz. 

\subsection{Point robot experiments} \label{subsec:point_robot}

First, we present the results for the Point robot. We model the Point robot using an \emph{extended kinematic unicycle model}:

\begin{equation} \label{eqn:point_model}
    \begin{bmatrix}
        \dot{x}_p \\
        \dot{y}_p \\
        \dot{\theta}_p \\
        \dot{v}_x \\
        \dot{v}_y \\
        \dot{\omega} \\
    \end{bmatrix}
    = \begin{bmatrix}
        v_p \cos (\theta_p) \\
        v_p \sin (\theta_p) \\
        K_\omega c_\omega\\
        K_{v2} (K_{v1} c_v - v_x) \\
        0 \\
        0 \\
    \end{bmatrix}
\end{equation}
where the state $\bm{x} = [x_p, y_p, \theta_p, v_x, v_y, \omega]^T$, the control $\bm{u} = [v_t \ \omega]^T$, $(x_p, y_p)$ is the center position of the robot, $\theta_p$ is the yaw angle, $v_x,v_y$ are the body frame longitudinal and lateral velocities and $\omega$ is the angular velocity. $c_v,c_\omega$ are the target velocity and angular velocity commands. Note that we have lateral and angular velocities included as states but they are not used in the kinematic model. This is just to enable the low-fidelity kinematic model to have consistent dimension with the high-fidelity dynamic model, thus facilitating the residual model learning algorithm \cite{Kalaria2023AdaptivePA}.

\begin{figure}
        \centering
        \includegraphics[width=.3\textwidth]{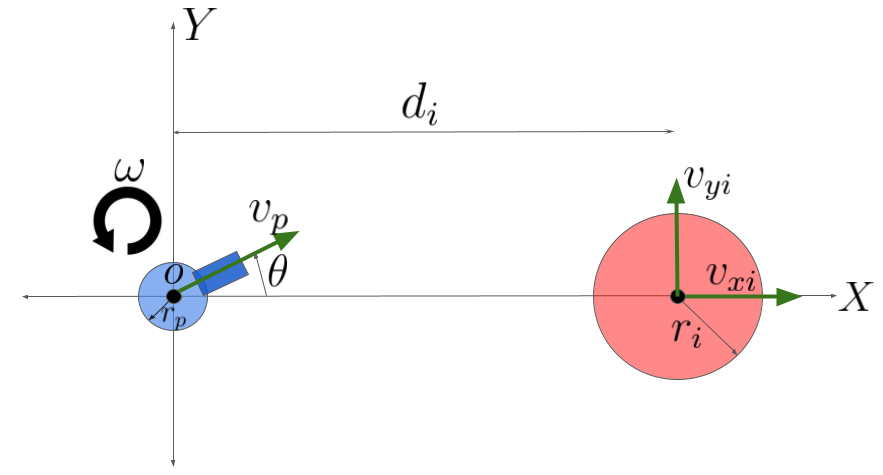}
        \caption{Reference frame for obstacle avoidance. Blue circle: ego robot; orange circle: moving obstacle.}
        \label{fig:point_ref}
\end{figure}

We formulate the safety specification of a robot to avoid any $i$-th circular region in space that is moving with a constant velocity.  We assume that any moving obstacle moves with a constant velocity $v_{xi}$ and $v_{yi}$ with the frame of reference visualized in Fig. \ref{fig:point_ref}. Let the radius of the moving obstacle be $r_i$ and of the robot be $r_p$. Without loss of generality, we derive the CBF for a frame of reference with origin at the robot and the $X$ axis direction along the circular region to be avoided. Let the angle made with the $X$ axis (i.e., the direction towards the obstacle) be $\theta$. Let the distance between the obstacle and the robot be $d_i$. To avoid a collision, we need to have $d - r_i - r_p \ge 0$. Therefore, we define $h_i(X) = d_i - r_i - r_p = \sqrt{(x_p-x_{ci})^2+(y_p-y_{ci})^2}$.


\begin{figure}[htbp]
\centering
\begin{subfigure}{.23\textwidth}
    \centering
    \includegraphics[width=\textwidth]{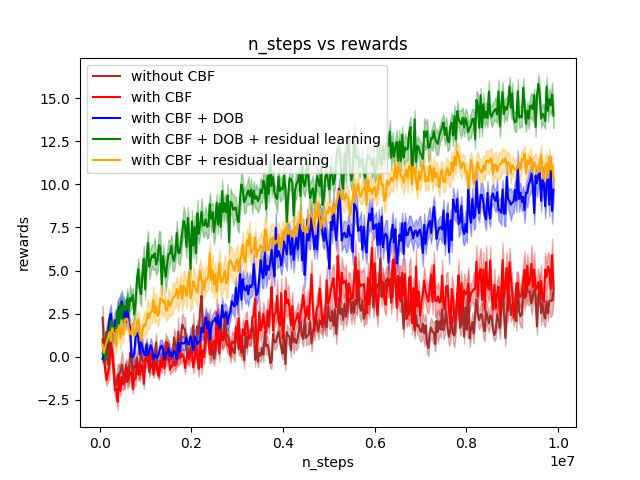}
    \caption{Rewards}
    \label{fig:rewards}
\end{subfigure}
\begin{subfigure}{.23\textwidth}
    \centering
    \includegraphics[width=\textwidth]{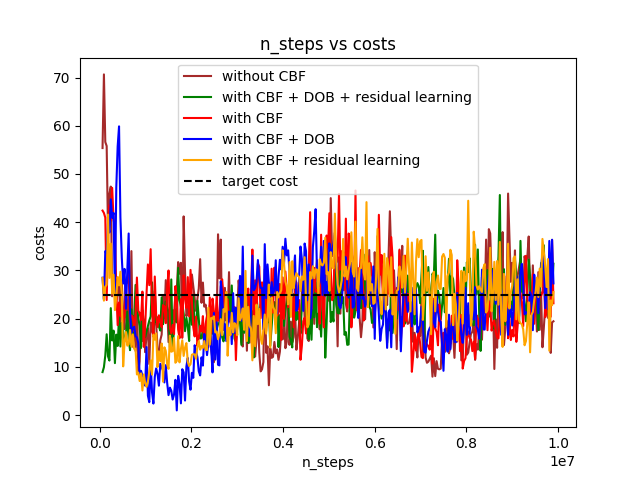}
    \caption{Costs}
    \label{fig:costs}
\end{subfigure}
\caption{Comparisons for Goal1 task on Point robot}
\label{fig:goal1}
\end{figure}

\begin{figure}[htbp]
\centering
\begin{subfigure}{.23\textwidth}
    \centering
    \includegraphics[width=\textwidth]{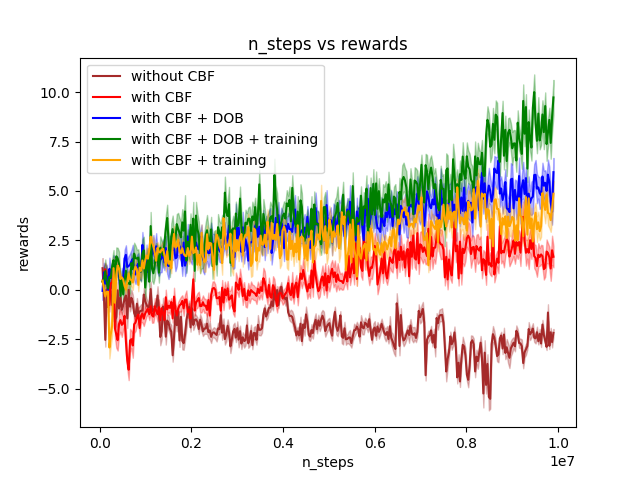}
    \caption{Rewards}
    \label{fig:rewards2}
\end{subfigure}
\begin{subfigure}{.23\textwidth}
    \centering
    \includegraphics[width=\textwidth]{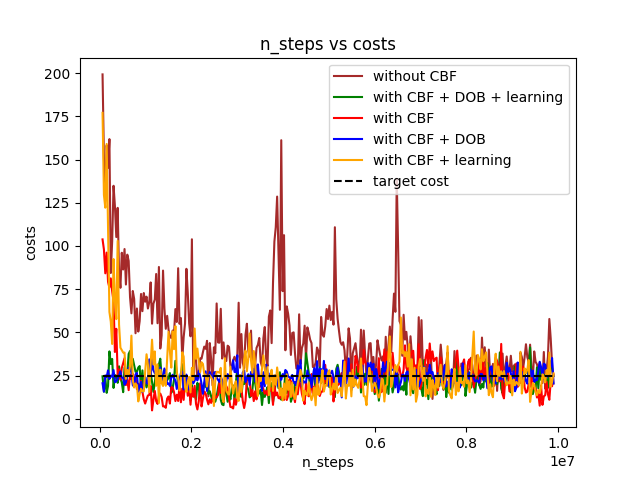}
    \caption{Costs}
    \label{fig:costs2}
\end{subfigure}
\label{fig:goal2}
\caption{Comparisons for Goal2 task on Point robot}
\end{figure}

\begin{figure}[htbp]
\centering
\begin{subfigure}{.23\textwidth}
    \centering
    \includegraphics[width=\textwidth]{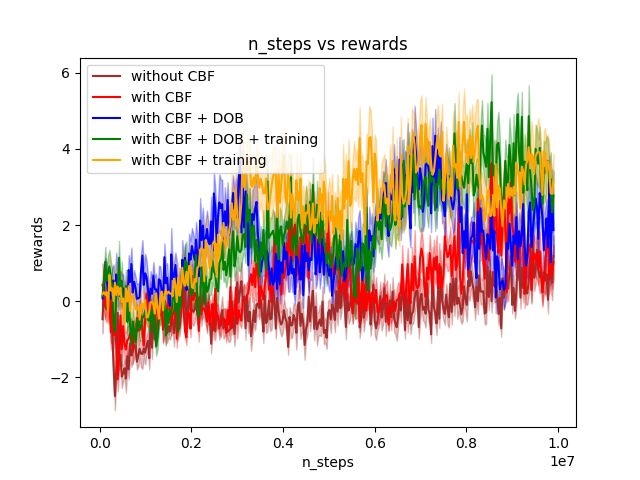}
    \caption{Rewards}
    \label{fig:rewards3}
\end{subfigure}
\begin{subfigure}{.23\textwidth}
    \centering
    \includegraphics[width=\textwidth]{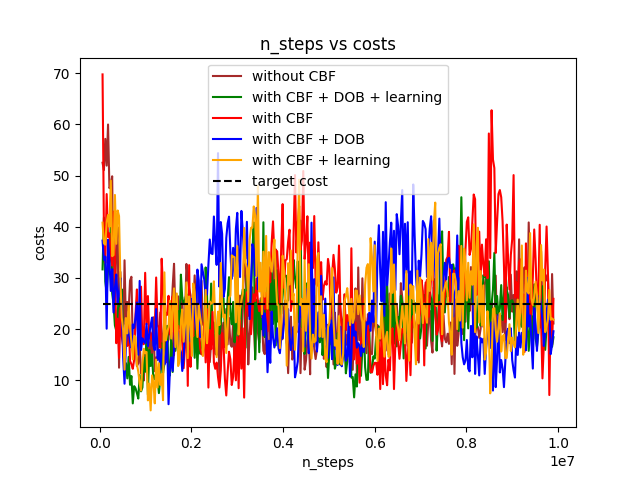}
    \caption{Costs}
    \label{fig:costs3}
\end{subfigure}
\label{fig:button1}
\caption{Comparisons for Button1 task on Point robot}
\end{figure}

\begin{figure}[htbp]
\centering
\begin{subfigure}{.23\textwidth}
    \centering
    \includegraphics[width=\textwidth]{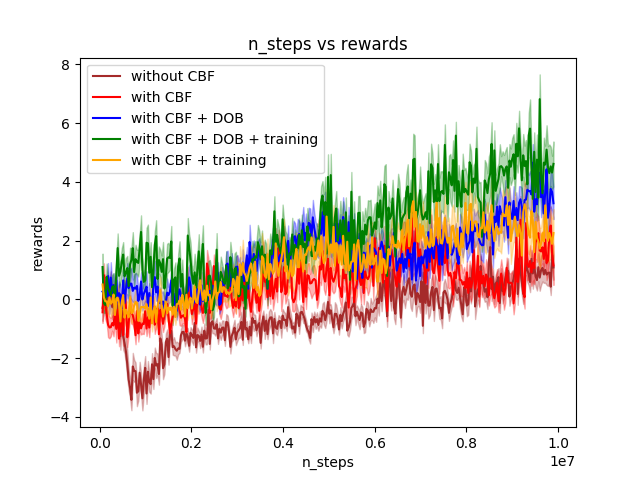}
    \caption{Rewards}
    \label{fig:rewards4}
\end{subfigure}
\begin{subfigure}{.23\textwidth}
    \centering
    \includegraphics[width=\textwidth]{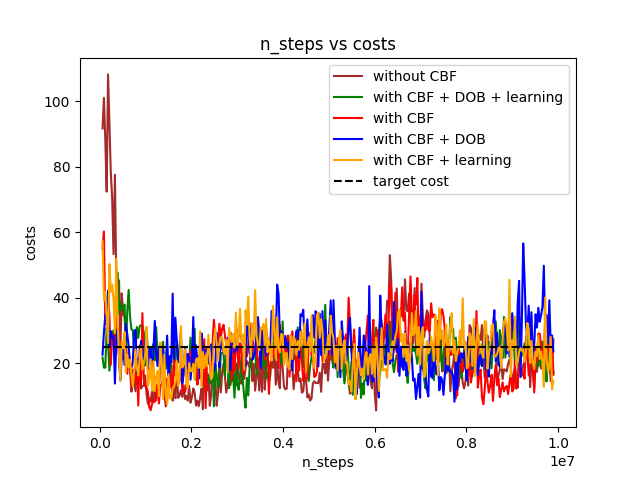}
    \caption{Costs}
    \label{fig:costs4}
\end{subfigure}
\label{fig:button2}
\caption{Comparisons for Button2 task on Point robot}
\end{figure}

\begin{figure}[htbp]
\centering
\begin{subfigure}{.23\textwidth}
    \centering
    \includegraphics[width=\textwidth]{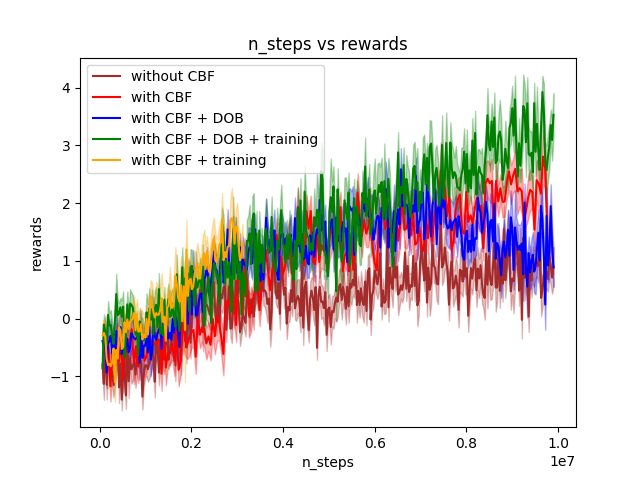}
    \caption{Rewards}
    \label{fig:rewards5}
\end{subfigure}
\begin{subfigure}{.23\textwidth}
    \centering
    \includegraphics[width=\textwidth]{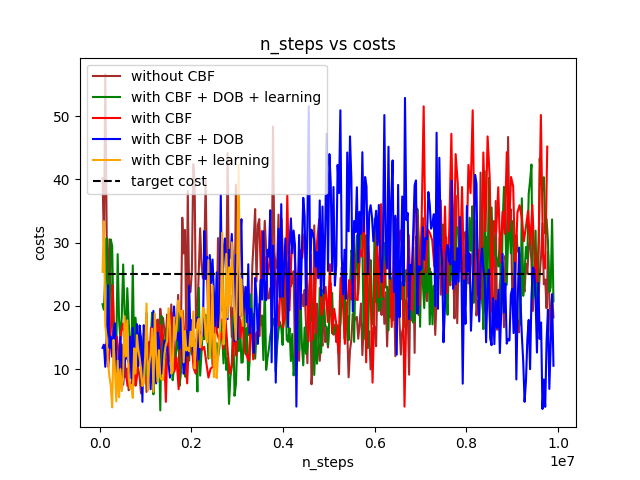}
    \caption{Costs}
    \label{fig:costs5}
\end{subfigure}
\label{fig:push1}
\caption{Comparisons for Push1 task on Point robot}
\end{figure}

\begin{figure}[htbp]
\centering
\begin{subfigure}{.23\textwidth}
    \centering
    \includegraphics[width=\textwidth]{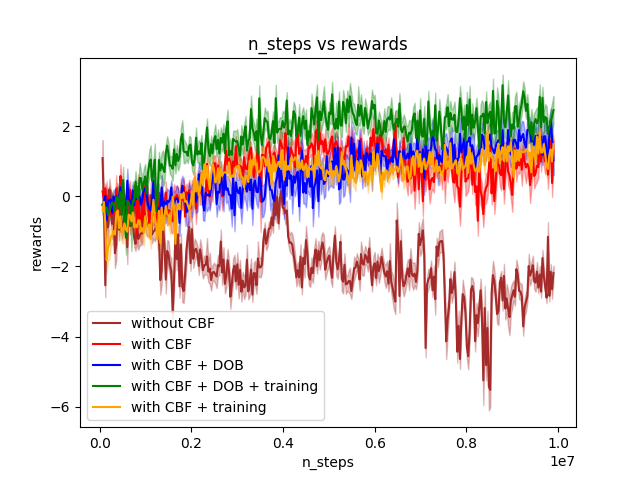}
    \caption{Rewards}
    \label{fig:rewards6}
\end{subfigure}
\begin{subfigure}{.23\textwidth}
    \centering
    \includegraphics[width=\textwidth]{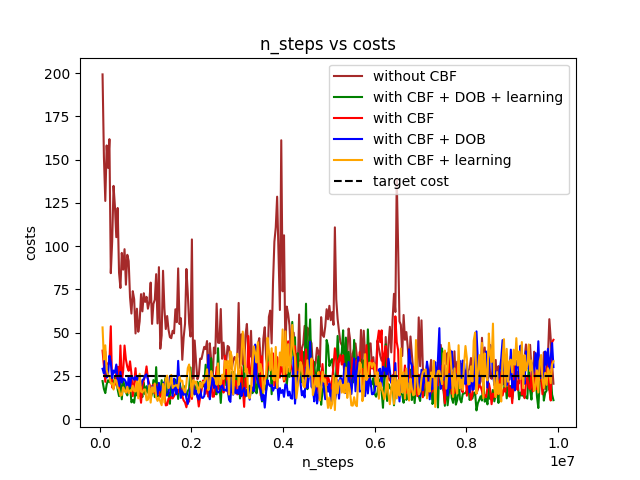}
    \caption{Costs}
    \label{fig:costs6}
\end{subfigure}
\caption{Comparisons for Push2 task on Point robot}
\label{fig:push2}
\end{figure}

The Point robot results in all environments are given in Figs. \ref{fig:goal1}-\ref{fig:push2}. Without CBF and in the presence of external disturbance the RL algorithm is not able to learn anything useful towards the task while remaining safe, as the reward is negative (see brown lines). Having CBF certainly improves, but due to external disturbances and model error, it fails to learn an efficient policy (see red lines). Adding DOB or residual model learning results in better rewards, but using only one of them still lags in achieving an agile policy (see blue and yellow lines). Our approach using both residual model learning and DOB is able to achieve the maximum reward in all environments (see green lines).

\subsection{Car robot experiments} \label{subsec:car_robot}

\begin{figure}[htbp]
\centering
\begin{subfigure}{.23\textwidth}
    \centering
    \includegraphics[width=\textwidth]{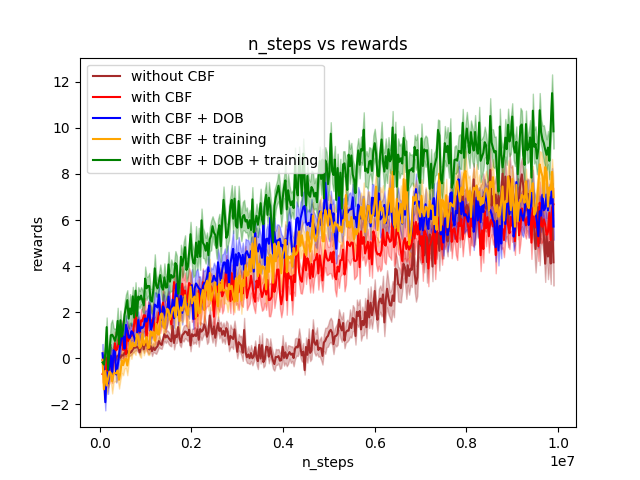}
    \caption{Rewards}
    \label{fig:rewards1_car}
\end{subfigure}
\begin{subfigure}{.23\textwidth}
    \centering
    \includegraphics[width=\textwidth]{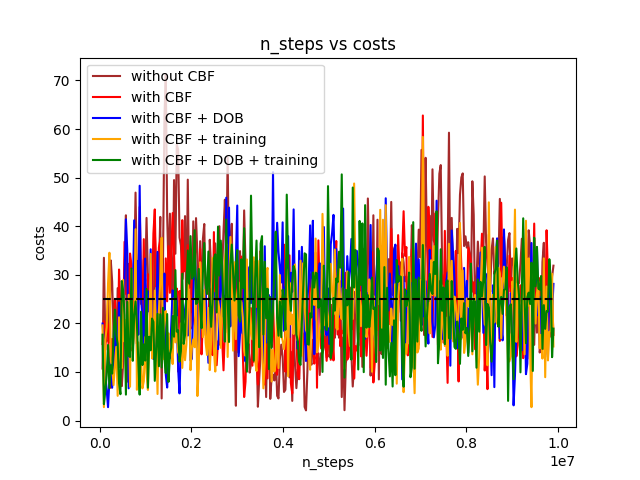}
    \caption{Costs}
    \label{fig:costs1_car}
\end{subfigure}
\caption{Comparisons for Goal1 task on Car robot}
\label{fig:car_goal1}
\end{figure}

\begin{figure}[htbp]
\centering
\begin{subfigure}{.23\textwidth}
    \centering
    \includegraphics[width=\textwidth]{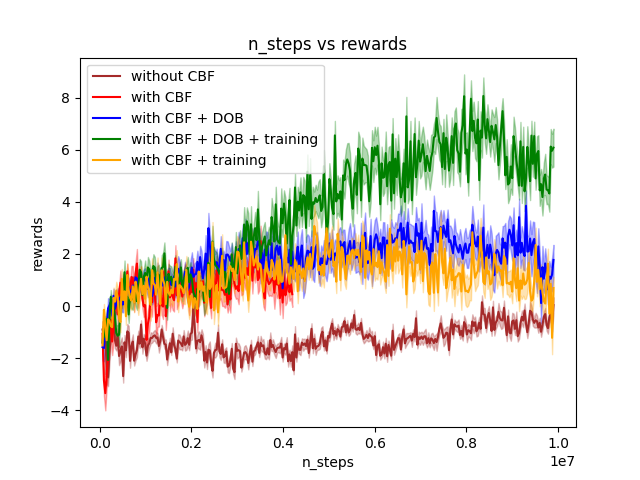}
    \caption{Rewards}
    \label{fig:rewards2_car}
\end{subfigure}
\begin{subfigure}{.23\textwidth}
    \centering
    \includegraphics[width=\textwidth]{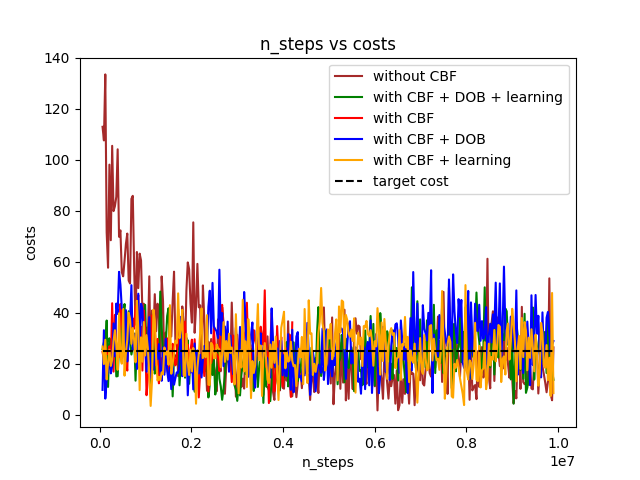}
    \caption{Costs}
    \label{fig:costs2_car}
\end{subfigure}
\caption{Comparisons for Goal2 task on Car robot}
\label{fig:car_goal2}
\end{figure}

\begin{figure}[htbp]
\centering
\begin{subfigure}{.23\textwidth}
    \centering
    \includegraphics[width=\textwidth]{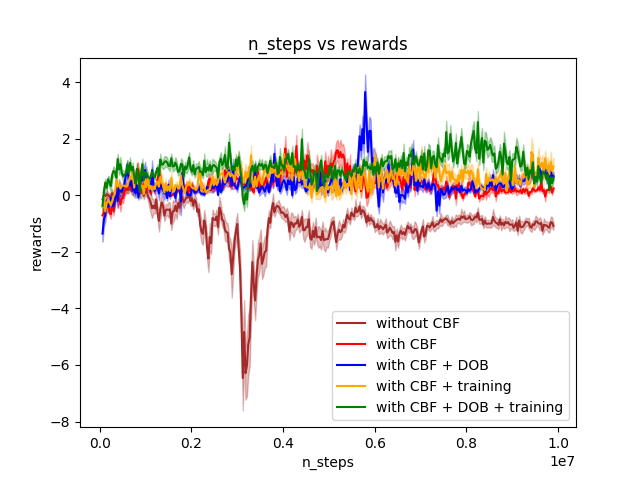}
    \caption{Rewards}
    \label{fig:rewards3_car}
\end{subfigure}
\begin{subfigure}{.23\textwidth}
    \centering
    \includegraphics[width=\textwidth]{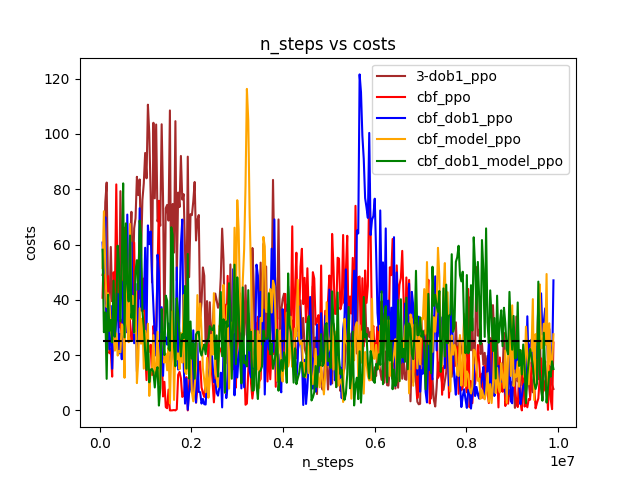}
    \caption{Costs}
    \label{fig:costs3_car}
\end{subfigure}
\caption{Comparisons for Button1 task on Car robot}
\label{fig:car_button1}
\end{figure}

\begin{figure}[htbp]
\centering
\begin{subfigure}{.23\textwidth}
    \centering
    \includegraphics[width=\textwidth]{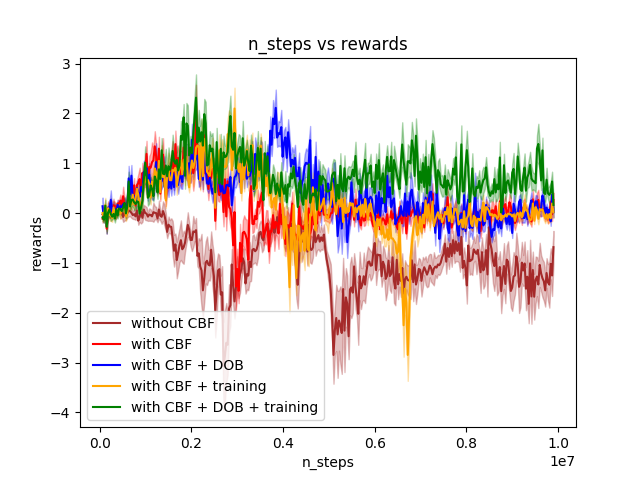}
    \caption{Rewards}
    \label{fig:rewards4_car}
\end{subfigure}
\begin{subfigure}{.23\textwidth}
    \centering
    \includegraphics[width=\textwidth]{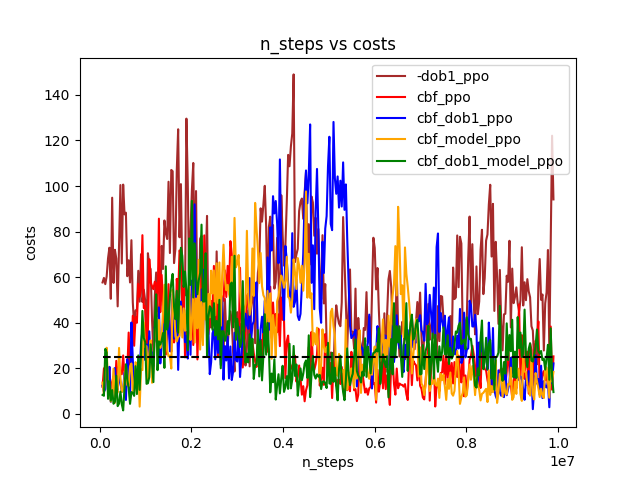}
    \caption{Costs}
    \label{fig:costs4_car}
\end{subfigure}
\caption{Comparisons for Button2 task on Car robot}
\label{fig:car_button2}
\end{figure}

\begin{figure}[htbp]
\centering
\begin{subfigure}{.23\textwidth}
    \centering
    \includegraphics[width=\textwidth]{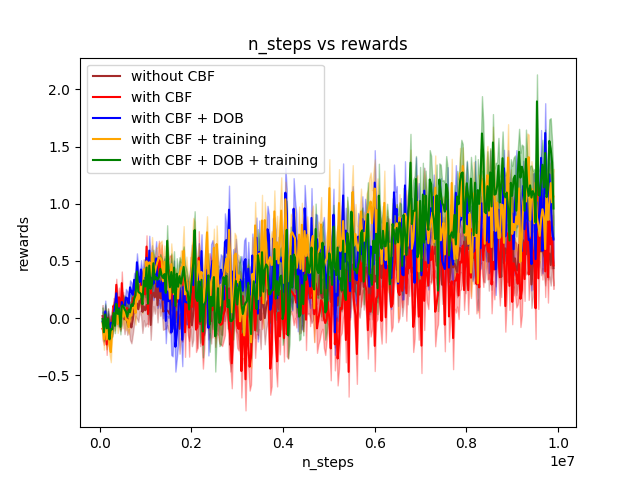}
    \caption{Rewards}
    \label{fig:rewards5_car}
\end{subfigure}
\begin{subfigure}{.23\textwidth}
    \centering
    \includegraphics[width=\textwidth]{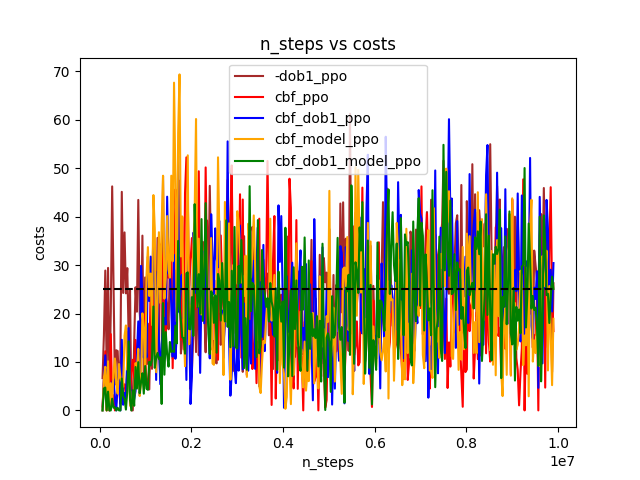}
    \caption{Costs}
    \label{fig:costs5_car}
\end{subfigure}
\caption{Comparisons for Push1 task on Car robot}
\label{fig:car_push1}
\end{figure}

For a Car robot, the control commands are the left and right wheel angular velocities, $c_l$ and $c_r$ respectively. For the nominal kinematic model, we simply change $\dot{v}_x = K_{v1} ( K_{v2} (c_l + c_r) - v_x)$ and $\dot{\omega} = K_\omega (c_r - c_l)$. $K_{v1}$, $K_{v2}$ and $K_\omega$ are approximated offline. Learning an optimal policy for the Car robot is more challenging, as the left and right wheel commands need to be coordinated. Also, there is more difference between the actual dynamic model and the kinematic model for the Car robot than the Point robot, meaning there is more skidding for the Car robot. Same as with point robot experiments, our approach is able to achieve maximum reward as compared to other baselines with the same cost constraints, see Figs. \ref{fig:car_goal1}-\ref{fig:car_push1}. 








\subsection{Physical RC car validations}

We also validated our framework on a physical F1/Tenth RC car. We design a simple task of going back and forth between two points similar to the Goal1 task of safety-gym but with only $1$ hazard in the middle and also with the added constraint of staying within the square arena of 4.2m $*$ 4.2m. Hence, we have CBF constraints for the $4$ walls and for the hazard at the center of radius 0.8 m with the same formulation as seen earlier in safety-gym experiments.

\begin{figure}[htbp]
    \centering
    \includegraphics[width=0.6\linewidth]{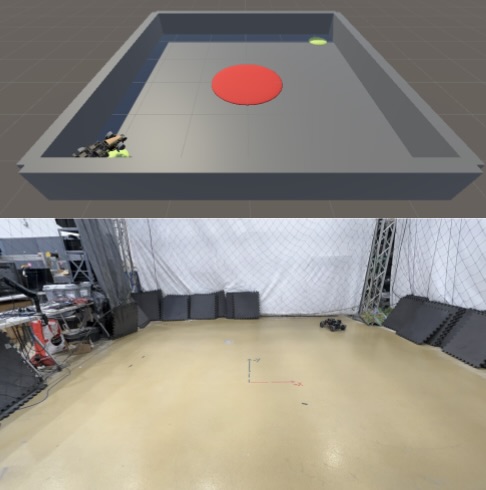}
    \caption{Snapshot of RC car simulator in Unity. The two yellow points are the positions the car tries to reach while the red zone is the hazard area.}
    \label{fig:rc_car_saferl_unity}
    \vspace{-7mm}
\end{figure}

Given limited battery endurance of the RC car, we are not able to train the RL completely from scratch on the physical testbed. Instead, we finish a pre-training of an initial policy in the Unity simulator (see Fig. \ref{fig:rc_car_saferl_unity}) with the same dimensions as the physical arena but with the only objective to reach the goal while not hitting the walls but not caring about the hazard. The pre-trained model on the simulator is then transferred to the hardware to continue training, with the added task of avoiding the hazard and trying to minimize the commute time between the two points. We test the same baselines as in safety-gym on this setup. \textbf{Note that the pre-training in simulation does not depart from our research goal of safe RL directly during the training, since collision avoidance is not even considered in the simulation training part. The most important training is still done with the hardware testbed.}

\begin{figure}[htbp]
\centering
\begin{subfigure}{.23\textwidth}
    \centering
    \includegraphics[width=\textwidth]{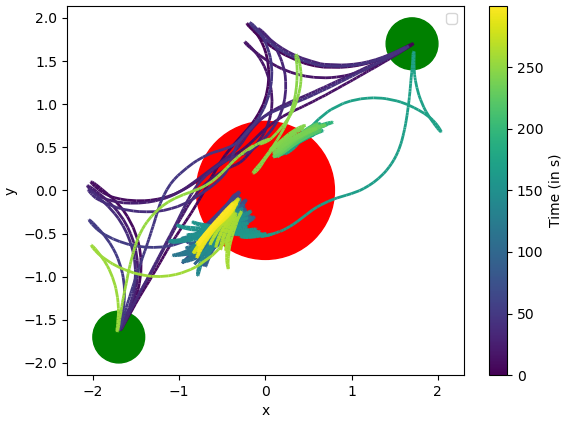}
    \caption{Without CBF}
    \label{fig:rc_saferl_wocbf}
\end{subfigure}
\begin{subfigure}{.23\textwidth}
    \centering
    \includegraphics[width=\textwidth]{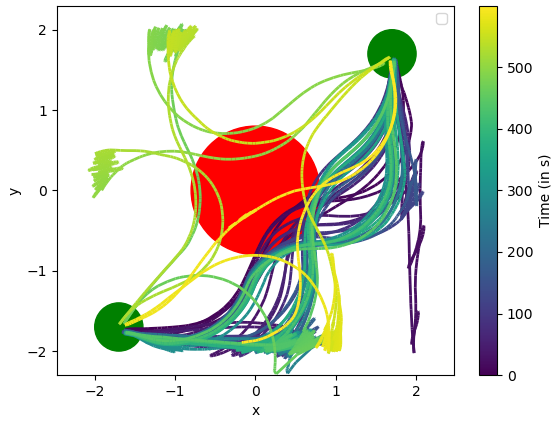}
    \caption{With DOB-CBF}
    \label{fig:rc_saferl_dob_cbf}
\end{subfigure}
\begin{subfigure}{.23\textwidth}
    \centering
    \includegraphics[width=\textwidth]{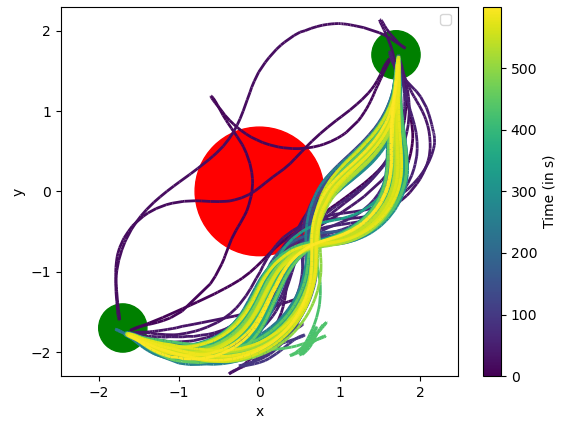}
    \caption{With RES-CBF}
    \label{fig:rc_saferl_res_cbf}
\end{subfigure}
\begin{subfigure}{.23\textwidth}
    \centering
    \includegraphics[width=\textwidth]{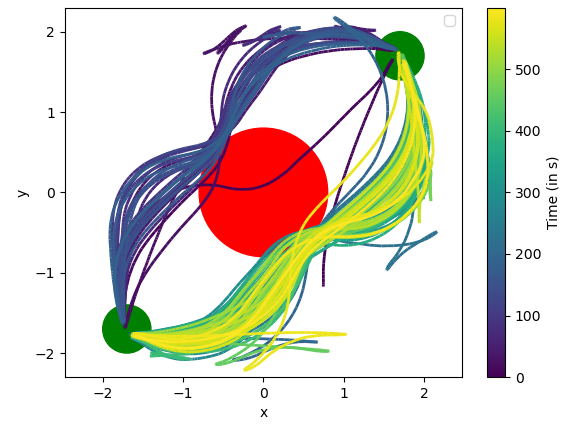}
    \caption{With RES-DOB-CBF}
    \label{fig:rc_saferl_res_dob_cbf}
\end{subfigure}
\caption{Comparisons of trajectories for different approaches in the RC car experiment}
\label{fig:rc_saferl}
\vspace{-2mm}
\end{figure}

\begin{figure}[H]
\centering
\begin{subfigure}{.23\textwidth}
    \centering
    \includegraphics[width=\textwidth]{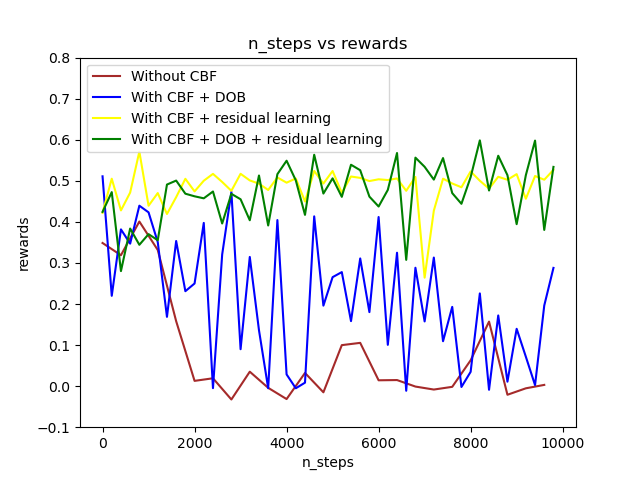}
    \caption{Rewards}
    \label{fig:rewards_rc}
\end{subfigure}
\begin{subfigure}{.23\textwidth}
    \centering
    \includegraphics[width=\textwidth]{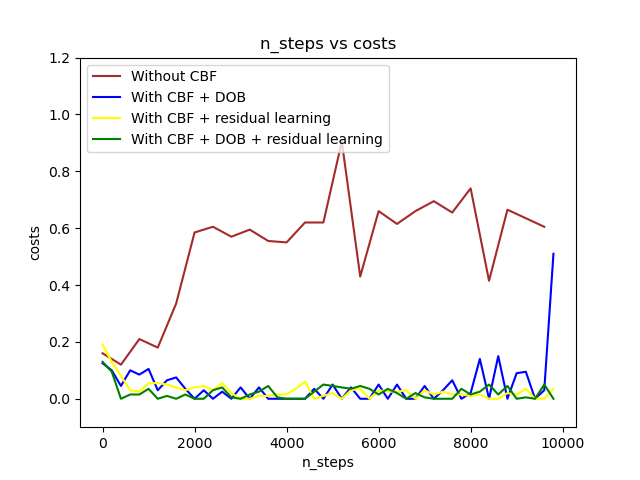}
    \caption{Costs}
    \label{fig:costs_rc}
\end{subfigure}
\caption{Comparisons of rewards and costs in the RC car experiment}
\label{fig:rc_reward_plot}
\end{figure}

We ran all baselines for 10 minutes on the hardware. We set the violation goal to 0 for PPO\_lagrangian. The trajectories for all baselines are shown in Fig. \ref{fig:rc_saferl}. The color bars indicate training progress, with the lightest color representing the approach to the end of the training time. The average commute times for all approaches during the last 5 minutes are listed in Table \ref{tab:rc_car_saferl_stats}. As shown in Fig. \ref{fig:rc_saferl_wocbf}, without CBF, the RC car struggles to learn a safe policy and ends with an average commute time 5.42 s. With \emph{DOB-CBF}, \emph{RES-CBF}, and \emph{RES-DOB-CBF},  it achieves good performance while avoiding hazards, with \emph{RES-DOB-CBF} achieving the best average time. The average number of safety violations is the average number of time steps during which the safety constraint was violated per trip from one goal to the other. The reward and cost comparisons throughout the run are shown in Fig. \ref{fig:rc_reward_plot}, which also demonstrates that our method achieves the minimum cost and maximum rewards.

\setlength{\tabcolsep}{5pt}
\begin{table}[]
\begin{tabular}{|l|l|l|l|l|}
\hline
                  & w/o CBF & RES-CBF \cite{Munoz2022} & DOB-CBF \cite{Cheng2022SafeAE} & Ours \\ \hline
\begin{tabular}[c]{@{}l@{}}Avg. comm- \\ ute time\end{tabular} & 5.42s       & 3.42s   & 3.47s   & \textbf{3.37s}       \\ \hline
\begin{tabular}[c]{@{}l@{}}Avg. no of \\ safety violations\end{tabular} & 146.1      & 7.33  & 2.60   & \textbf{2.07}       \\ \hline
\end{tabular}
\caption{Comparisons for average commute times (efficiency) and average number of safety violations (safety)}
\label{tab:rc_car_saferl_stats}
\end{table}

\section{Conclusion and Future Work} \label{sec:conclusion}
We introduce a disturbance rejection-guarded learning component that combines a disturbance observer with residual model learning. This component can be easily integrated with existing RL frameworks to provide robust and safe learning that addresses both internal and external disturbances. We have tested our approach on various scenarios within the Safety-Gym benchmark, as well as on a physical RC car. The results demonstrate that we achieve superior safety performance without compromising efficiency. 

\bibliographystyle{IEEEtran}
\bibliography{./IEEEfull,refs}


\end{document}